\documentclass[journal,twoside,web]{ieeecolor}
\usepackage{generic}
\usepackage{cite}
\usepackage{amsmath,amssymb,amsfonts}

\usepackage{graphicx}
\usepackage{makecell}
\usepackage{algpseudocode}
\usepackage{hyperref}

\usepackage{amsmath, amssymb, amsthm} 
\usepackage{amsmath} 
\usepackage{booktabs}
\usepackage{paralist}
\usepackage{multirow}
\usepackage{graphicx}
\usepackage{bbm}
\usepackage{booktabs}
\usepackage{array}
\usepackage{courier}   
\usepackage[T1]{fontenc}
\usepackage{booktabs}
\usepackage{multirow}
\usepackage{makecell}
\usepackage[table,xcdraw]{xcolor}
\definecolor{gray}{gray}{0.5}
\usepackage{amsmath}
\let\labelindent\relax
\usepackage[inline]{enumitem}

\theoremstyle{definition}

\hypersetup{hidelinks=true}
\usepackage{textcomp}
\usepackage{cite}
\usepackage{algorithm}
\usepackage{algpseudocode}
\usepackage{hyperref}  
\usepackage{tikz}

\newcommand\submittedtext{%
  \footnotesize This work has been submitted to the IEEE for possible publication. Copyright may be transferred without notice, after which this version may no longer be accessible.}

\newcommand\submittednotice{%
\begin{tikzpicture}[remember picture,overlay]
\node[anchor=north,yshift=-12pt] at (current page.north) {\parbox{\textwidth}{\centering\submittedtext}};
\end{tikzpicture}%
}
\providecommand{\refname}{References}   
\begin{document}
\title{Geometric Topology  for Multi-Site Uncertainty Aware Prediction of Body Mass Index from Structural MRI}
\author{\mbox{Walia Farzana, \IEEEmembership{Member, IEEE}}, Norou Diawara,
Liangsuo Ma,
F. Gerard Moeller, \\  \makebox[\textwidth][c]{Khan M. Iftekharuddin, \IEEEmembership{Senior Member, IEEE}}
\thanks{This work was supported in part by the NIH Grant \#R01 EB020683-06 and CTSA Grant \#UM1TR004360. Walia Farzana and Khan M. Iftekharuddin are with the Vision Lab,
Department of Electrical \& Computer Engineering, Old Dominion University,
Norfolk, VA 23529 USA (e-mail: wfarz001@odu.edu; kiftekha@odu.edu).
Norou Diawara is with the Department of Mathematics \& Statistics, Old
Dominion University, Norfolk, VA 23529 USA (e-mail: ndiawara@odu.edu).
Liangsuo Ma is with Institute of Drug and Alcohol Studies and F.Gerard Moeller is with the Department of Psychiatry, Virginia Commonwealth University, Richmond, VA 23284 USA(email:liangsuo.ma@vcuhealth.org, frederick.moeller@vcuhealth.org).}}

\maketitle
\submittednotice
\begin{abstract}
Quantitative neuroimaging methods, such as structural magnetic resonance imaging (sMRI), aim to extract anatomical phenotypes by estimating volume, cortical thickness, and topological measures (mean and Gaussian curvature) from gray and white matter. However, these scalar measures do not capture the direction of cortical folding patterns in the brain, and the corresponding methods lack uncertainty quantification body mass index (BMI) prediction in patient data from multi-site, limiting the reliability and usability needed for clinical translation. Cortical folding is related to brain functional organization, but quantifying its 3D geometry requires mathematical tools beyond volume and thickness.Consequently, the proposed computational framework in this work makes two novel contributions. First, we introduce geometry-based directional localized Darboux frame features (DFF) from 3D sMRI meshes that encode the direction of cortical folding, which scalar measures miss. Second, we propose a dynamic fusion layer that integrates disentangled semantic representations (DSR) from multi-site sMRI and DFF followed by multi-site uncertainty quantification. We evaluate the proposed framework on data from three independent sites, with a total of 1205 subjects and an imbalanced distribution across BMI groups. Our method outperforms re-weighting loss baselines, conventional features in performance comparison, and leave-one-site-out uncertainty quantification. In addition, we identify specific brain regions across all sites correlating with the BMI group, along with their associated uncertainty.
\end{abstract}

\begin{IEEEkeywords}
Structural MRI, Darboux-Frame, Geometric Topology, BMI, Uncertainty Quantification.
\end{IEEEkeywords}

\section{Introduction}
\label{sec:introduction}
Obesity is a global challenge shaped by biological, physiological, behavioral, social, environmental, and economic factors. By 2030, about 3.3 billion people (60\% of the global population) may be overweight or obese\cite{bib1}.BMI is a commonly used index of obesity, measuring excess body weight relative to height and age. People are classified as underweight (BMI $\leq$ 18.5), healthy weight (18.5 $\leq$ BMI $\le$ 25), overweight (25 $\leq$ BMI $\le$ 30), or obese (BMI $\geq$30) according to their BMI\cite{bib2}. Obesity affects all ages and backgrounds, increasing the risk of type 2 diabetes, cardiovascular disease, cancer, and dementia, including Alzheimer's \cite{bib3}.
Neuroimaging techniques, including functional and sMRI and diffusion tensor imaging (DTI), are increasingly used to investigate the neurobiological basis of obesity-related eating behaviors in humans\cite{bib4}. sMRI provides anatomical phenotypes, including gray matter volumes (GMV) and white matter (WM), and cortical morphometry. Most GM-focused studies report atrophy in frontal, temporal, and parietal regions, though findings vary substantially\cite{bib5}. Research also links BMI to changes in brain morphology and microstructure \cite{bib6}.
Thus, BMI-related white matter microstructural changes mediate cortical changes, measured by brain-surface Gaussian curvature \cite{bib7}. Curvature metrics \cite{bib8} examine cortical morphology through convolution patterns.

\begin{figure}[h]
    \centering
    \includegraphics[width=0.28\textwidth]{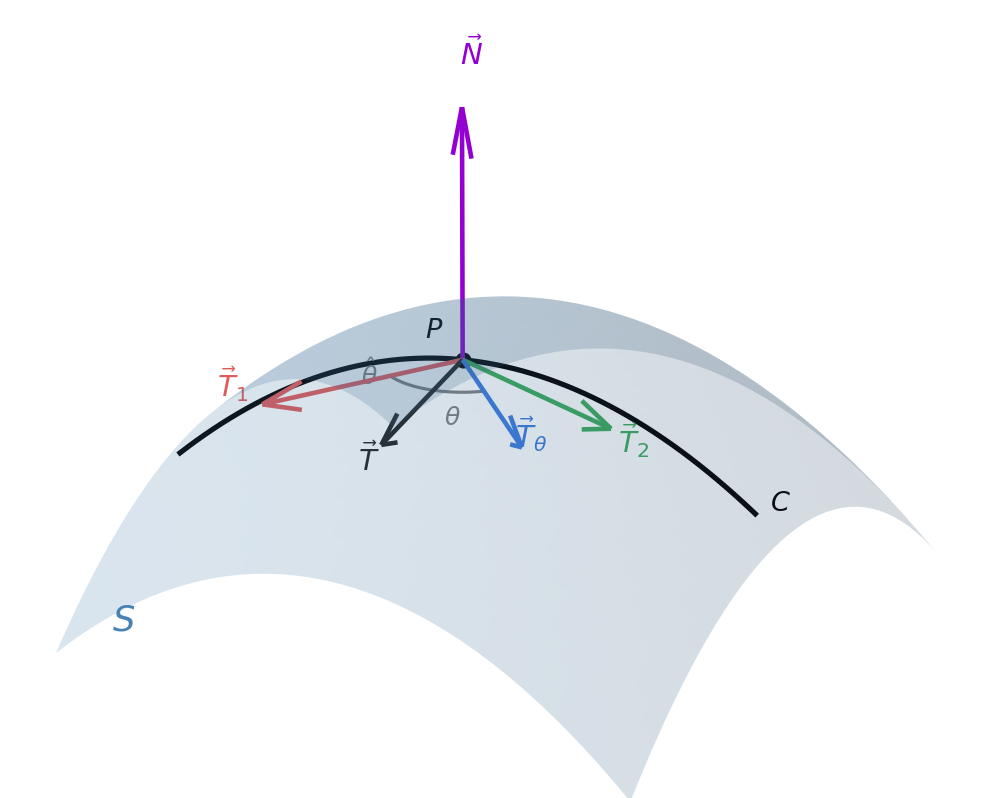}
    \caption{Surface $S$: curve $C$, normal $\vec{N}$, and tangent directions $\vec{T_1}, \vec{T}, \vec{T_{\theta}}, \vec{T_2}$ at $P$.}
    \label{fig:my_figure}
\end{figure}
The cerebral cortex has ridges called gyri and valleys called sulci.These folds occur not only in the radial direction, forming gyri and sulci, but also in the tangential direction\cite{bib9}.
At a point on a curved surface, two orthogonal planes correspond to maximum and minimum principal curvatures, $k_1$ and $k_2$.  These principal curvatures determine the local shape of the surface point \cite{bib8}. In addition to the principal curvatures at each surface point, the local coordinate system includes the surface normal $\vec{N}$ and the two principal tangent directions $\vec{T_1},\vec{T_2}$, as shown in Fig.\ref{fig:my_figure} forming the Darboux frame \cite{bib10}. \\
Existing  \cite{bib6,bib7} BMI-related cortical topological measures (mean curvature, Gaussian curvature) reflect scalar values of global cortical folding but do not specify the directional geometry of the folding pattern. \\
A second challenge is that findings on BMI and brain structure remain inconsistent. Most studies focus on GMV and have found decreased total GMV in obesity, while some report no reductions \cite {bib11}. The exact localization of these changes varies widely, and such inconsistencies make it challenging to understand the association between weight dysregulation and the brain. 
There are three likely causes for these inconsistencies.\\
First, limited sample size and lack of multi-site data cause significant variability\cite{bib5}. `Site' refers to a single MRI scanner protocol. Incorporating multi-site data is essential to create robust models that can function across various healthcare institutions \cite{bib24}. \\
Second, comorbidities that are independently associated with brain atrophy, such as psychiatric disorders or alcohol consumption, act as confounders, causing inconsistent findings across different multi-site dataset \cite{bib11}. Third, methods used to examine brain structure vary in their sensitivity to different structural changes. Most studies focus on voxel-based morphology \cite{bib11,bib12,bib13,bib14,bib15}, with a few on scalar quantification of cortical folding on WM surface\cite{bib7}. These approaches rarely quantify predictive uncertainty, limiting their reliability for clinical acceptance.\\ 
To address these challenges, we make two novel contributions. First, we introduce DFF that encodes the geometric direction of cortical region folding. Second, we introduce a dynamic fusion layer for multi-site data, integrating DSR with DFF, followed by predictive multi-site uncertainty quantification.

\begin{figure*}[t] 
    \centering
    \includegraphics[width=0.75\textwidth]{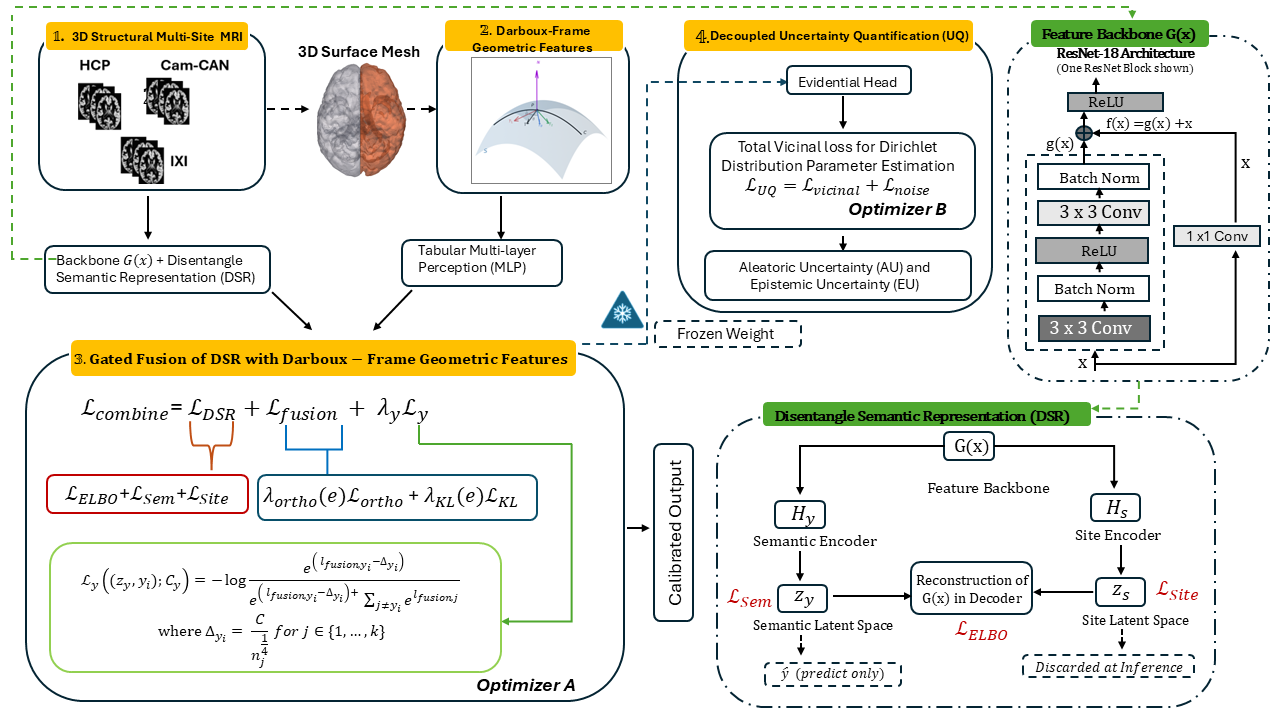}
    \caption{An overview of the proposed Darboux-Frame Based Geometric Representation Learning for Multi-Site Uncertainty Aware Prediction of Body Mass Index. The orange color blocks refers to the core components of the proposed method and green color blocks depicts feature backbone $G(x)$ within DSR and Disentangle Semantic Representation (DSR) respectively.}
    \label{fig:wide_figure}
\end{figure*}
\subsection{Related Work}
sMRI measures macroscopic changes in specific brain regions associated with obesity\cite{bib15}. Voxel-based morphometry measures white or gray matter volume; surface-based morphometry measures cortical thickness, area, and folding \cite{bib12}. 
Obese participants had reduced GMV in frontal regions (medial prefrontal and orbitofrontal cortices) and mesolimbic regions (anterior cingulate cortex, caudate, putamen, nucleus accumbens, insula, amygdala, hippocampus, and pallidum)\cite{bib4,bib13}. Reductions also occurred in the thalamus and anterior corpus callosum; all were negatively correlated with BMI.
BMI is negatively associated with WM volume (WMV) in the corpus callosum, thalamic radiations, capsules, fascicles, uncinate fasciculus, and cingulum \cite{bib14}. 
Region-based surface morphometry reveals reduced cortical thickness in the left lateral occipital cortex (LOC) and the right ventromedial prefrontal cortex (VmPFC), without a significant association with global surface area and folding\cite{bib11}. Familial obesity risk and current obesity/overweight are associated with overlapping and distinct brain changes, including reduced GMV, WMV, and cortical thickness in the postcentral gyrus (somatosensory cortex), opercular cortex (taste cortex), anterior cingulate cortex, and precuneus \cite{bib16}.\\
Evidence suggests obesity affects white matter microstructure, measured by diffusion tensor imaging (DTI), T1-weighted MRI hyperintensities (WMH), and millimeter-scale Gaussian curvature\cite{bib17,bib18}.
Overweight and obese individuals showed age-related reductions in DTI-measured fractional anisotropy (FA) in the right superior longitudinal fasciculus, corpus callosum splenium, and right cingulum. The frontoparietal network, mainly linked to cognitive control in individuals with obesity, involves the orbitofrontal cortex as a key region connecting WMH and obesity \cite {bib19,bib20}.Gaussian curvature of the WM surface, a sensitive T1-weighted MRI measure for tracking obesity effects on cortical morphology, was positively associated with BMI, mediated by a negative association between white matter FA and BMI\cite{bib7}. However, scalar curvature measures alone discard directional information and may miss BMI-related differences in folding geometry. \\
 Deep Learning (DL) based approaches, such as computational models based on   Convolutional Neural Networks (CNN)  \cite{bib21, bib22, bib23}, predict BMI from sMRI, however, these methods are trained and evaluated on a single site or private data. Hence, generalization of these methods to unseen multi-site data is unknown, and they provide no uncertainty estimates to indicate when a prediction is reliable.

\section{Darboux Frame Features from Cortical Surface Mesh}
Scalar curvature methods inherently fail to capture the direction of folding in cortical layers. Therefore, we introduce a Darboux frame to each point on the cortical surface. The computation requires formulating frames from a continuous surface to a discrete mesh and estimating normal and tangent directions from the discrete mesh. The following subsections provide a mesh-based formulation\cite{bib10}. Our contribution is region-specific directional features of cortical folding, which, to our knowledge, is the first application of Darboux frames in geometric direction for BMI prediction. 
 
Let $\vec{T}_1$,  $\vec{T}_2$ be  the principal directions at point P ( Fig.\ref{fig:my_figure}) with the corresponding principal curvatures ${k}_1$ and ${k}_2$ respectively. For every  direction with angle $\theta$ between $\vec{T}_\theta$ and $\vec{T}_1$, the directional curvature at point $P$

\begin{equation}\label{eq: directional curvarture1}
{k}_p(\vec{T}_\theta) = {k}_1cos^{2}\theta + {k}_2sin^{2}\theta.
\end{equation}

We inversely measure ${k}_p$($\vec{T}_\theta$) in various directions $\theta$ to recover ${k}_1$ and ${k}_2$ and the principal directions.
\vspace{-0.3cm}
\subsection{Continuous to Discrete Case}
Consider C(s) as the arclength parameterized normal section at P along the direction $\vec{T}_\theta$.
According to the parameterization: C(0)=P, $C'$(0)= $\vec{T}_\theta$, $C''$(0)= ${k}_p$($\vec{T}_\theta$)$\vec{N}$. Expanding C(s) to a Taylor Series near $s=0$, we obtain
\begin{equation}\label{eq: Taylor exp}
C(s)- P = \vec{T}_\theta s + \frac{1}{2} {k}_p(\vec{T}_\theta)\vec{N}s^{2} + O(s^{3}).
\end{equation}
Now, multiplying both sides of Equation (\ref{eq: Taylor exp}) with 2$\vec{N}^{T}$, we obtain
2$\vec{N}^{T}$(C(s)- P) = 2$\vec{N}^{T}$$\vec{T}_\theta$s+ ${k}_p$($\vec{T}_\theta$)$s^{2}$ + $O(s^{3})$. Since the surface normal $\vec{N}$ is perpendicular to all tangent vectors $\vec{T}_\theta$, the first term vanishes, 2$\vec{N}^{T}$$\vec{T}_\theta$$=0$. Therefore,

\begin{equation}\label{eq: product}
2\vec{N}^{T} (C(s)- P) = {k}_p(\vec{T}_\theta) s^{2} + O(s^{3}).
\end{equation}

Taking the length of vectors on both sides of Equation (\ref{eq: product}) 
leads to
\begin{equation}\label{eq: norm}
 ||(C(s)- P)||^2 = s^{2} + O(s^{4}).
\end{equation}

By dividing the last Equations (\ref{eq: product}) with (\ref{eq: norm}), we have:
\begin{equation}
\frac{2\vec{N}^{T} (C(s)- P)} {||(C(s)- P)||^2} = {k}_p(\vec{T}_\theta) +O(s).
\end{equation}

Now if we take a point Q, lying on surface \textbf{S} very close to P and let $\vec{T}_\theta$ denote the projection of the vector $\overrightarrow{PQ}$ on the tangent plane to \textbf{S} at P, we can approximate ${k}_p$$(\vec{T}_\theta)$ in Equation (\ref{eq: directional curvarture1}) as 
\begin{equation}\label{eq: directional curvature2}
{k}_p(\vec{T}_\theta) \approx \frac{2\vec{N}^{T} (Q- P)} {||(Q- P)||^2} .
\end{equation} 
For each mesh vertex $V \in \mathbb{R}^3$ with local mesh-neighbor index $i$, where $i=1,2,....,n_i$, for vertex $V_{i}$ with projected tangent direction $\vec{T}_{i}$ and estimated surface normal $\vec{N_{V}}$ at vertex $V$, the directional curvature ${k}_i^{V}$

\begin{equation}
{k}_i^{V}(\vec{T}_{i}) \approx \frac{2\vec{N_{V}}^{T} (V_{i}- V)} {||(V_{i}- V)||^2}.
\end{equation}

\subsection{Computation of Surface Normal}
Computation of surface normal $\vec{N_{V}}$ at vertex $V$, using triangles formed by $V$ and pairs of immediately adjacent neighbors. For each adjacent triangle ($V$,$V_{i}$,$V_{i+1}$), compute the unit face normal ${n_{i,i+1}}$ and angle weight $w_{i,i+1}$.

For each such triangle ($V$,$V_{i}$,$V_{i+1}$)
\begin{equation}
\vec{n}_{i,i+1} = \frac {(V_{i} - V) \times (V_{i+1}-V)} {||(V_{i} - V) \times (V_{i+1}-V) || } ,
\end{equation} \\
where, $\times$ is the cross-product,
\begin{equation}
w_{i,i+1} = \arccos (\frac {(V_{i} - V) \cdot (V_{i+1}-V)} {||(V_{i} - V) \cdot (V_{i+1}-V) || } ) ,
\end{equation} \\
is the angle, and the estimated vertex normal is
\begin{equation}
\vec{N_{V}} = \frac {\sum_{i,i+1}w_{i,i+1}\vec{n}_{i,i+1}} {|| \sum_{i,i+1}w_{i,i+1}\vec{n}_{i,i+1}|| } ,
\end{equation}
where the sum is over all adjacent triangle pairs of $V$ \cite{bib25}. 

\subsection{Computation of Tangent Direction}
Tangent direction $\vec{T}_{i}$ is the projection of the vector connecting two neighboring vertices on the tangent plane to surface \textbf{S} at vertex $V$. Consider an extended neighborhood $\tilde{V}$ of all vertices with topological ring radius $r$ and width $w$ around $V$. For each neighbor $V_{i}$, compute tangent direction $\vec{T}_{i}$ by projecting the vector $(V_{i}-V)$ onto the tangent plane at $V$ by subtracting the component along $\vec{N_{V}}$.

\begin{equation}
\vec{T}_{i} = (V_{i}-V) - [ \vec{N_{V}}^{T}\cdot(V_{i}-V)]\vec{N_{V}} .
\end{equation}

\subsection{Algorithm for Darboux Frame Features}
The algorithm is focused on constructing a matrix $M_{P}$, expressed as an integral at each 3-D point where principal curvature and Darboux Frame need to be estimated. 
\subsubsection{Continuous to Discrete Case }
\textit{Step 01-Defining $M_{P}$}:
An arbitrary tangent $\vec{T}$ to surface \textbf{S} at point $P$. For $-\pi \leq \theta \leq \pi$, $\vec{T_{\theta}}$ denote the tangent to \textbf{S} at $P$ in the direction which creates an angle $\theta$ with $\vec{T}$.

\begin{equation}
M_{P} = \frac {1}{2\pi} \int_{-\pi}^{\pi} {k}_p(\vec{T}_\theta)\vec{T}_\theta\vec{T}_\theta^T d\theta .
\end{equation}

\textit{Step 02-Parameterization of  $\vec{T}_\theta$:}

Let, $\hat{\theta}$ is the angle between $\vec{T}$ and the unknown principal direction $\vec{T_{1}}$ as shown in Fig.1 , then

\begin{equation}
\vec{T}_\theta = \vec{T_{1}}cos(\theta +\hat{\theta}) + \vec{T_{2}}sin(\theta +\hat{\theta}) .
\end{equation}

\textit{Step 03-Principal Directions ($\vec{T_{1}}$, $\vec{T_{2}}$) Eigenvectors of $M_{P}$ }: Utilizing Euler's Formula (${k}_p$$(\vec{T}_\theta)$ = ${k}_1$$cos^{2}\theta$ + ${k}_2$$sin^{2}\theta$) and Equation (12) we can get $M_{P}\vec{T_{1}}$. 
Over $[-\pi,\pi]$, the cross-term terms with odd powers vanish because trigonometric integrals over a full period. Using $\int cos^{4} (\theta +\hat{\theta})=\frac{3\pi}{4}$ and $\int sin^{2}cos^{2} (\theta +\hat{\theta})=\frac{\pi}{4}$, the following can be shown

\begin{equation}
e_{1} =\vec{T}_{1}^{T}M_{P}\vec{T}_{1} = \frac{3}{8}k_{1} + \frac{1}{8}k_{2} .
\end{equation}

Similarly, $\vec{T_{2}}$ is eigenvector of $M_{P}$, therefore the following can be obtained
\begin{equation}
e_{2} =\vec{T}_{2}^{T}M_{P}\vec{T}_{2}  =\frac{1}{8}k_{1} + \frac{3}{8}k_{2} .
\end{equation}

\textit{Step 04-Eigendecomposition of matrix $M_{P}$:}

The surface normal $\vec{N}$ is always the eigenvector with eigenvalue $0$ because $\vec{N} \perp \vec{T}$ for all $\theta$ making integrand contribute to zero in the $\vec{N}$ direction. Therefore, any tangent direction $\vec{T}$ by which $M_{P}$ is constructed, the matrix can be factorized as the following

\begin{equation}
M_{P} = \begin{bmatrix} \vec{T}_{1} & \vec{T}_{2} & \vec{N} \end{bmatrix} 
\begin{bmatrix}
    e_{1} & 0 & 0 \\
    0 & e_{2} & 0 \\
    0 & 0 & 0
\end{bmatrix} 
\begin{bmatrix} \vec{T}_{1} & \vec{T}_{2} & \vec{N} \end{bmatrix}^{T} .
\end{equation}

T algorithm \cite{bib10} is the modification of Taubin's algorithm.
In discrete case, the surface \textbf{S} consists of vertices $V$, each one of the $V$ represents 3-D point on \textbf{S}. By connecting neighboring vertices the surface $S$ approximated as a polyhedral mesh.
Considering $\tilde{V}$ as the set of surface vertices containing the topological neighbors of vertices $V$ on polyhedral mesh. $\vec{T}_{i}$ represents the tangent to \textbf{S} at $V$ in the direction connected $V$ to $ {V}_{i} \in \tilde{V}$ (i.e. $\vec{T}_{i}$ is the projection of the vector connecting two vertices on the tangent plane to \textbf{S} at $V$, ${k}_i$ is the directional curvature associated with $\vec{T}_{i}$).

\textit{Step 01- Defining $M_{V}$ (Equivalence of $M_{P}$)}:

\begin{equation}
M_{V}=\sum_{{V}_{i} \in \tilde{V}} w_{i}k_{i}\vec{T}_{i}\vec{T}_{i}^{T} ,
\end{equation}

where $w_i$ is proportional to the geometric distance between $V$ and $V_i$, with $\sum_{{V}_{i} \in \tilde{V}}w_{i}=1$, so distant neighbors have less influence on the local approximation\cite{bib10}.
\begin{equation}
w_{i} \propto \frac{1}{|| V_{i} -V||} .
\end{equation}\\

\textit{Step 02- Computing Surface Normal, Tangent Direction and Directional Curvature:} For each vertex $ V _ {i} \in \tilde{V}$, compute the surface normal (Eq. 10), tangent direction (Eq.11), and directional curvature (Eq. 7).

\textit{Step 03- Eigendecomposition of $M_{V}$: }

Compute the eigenvalues and eigenvectors of $M_{V}$ using singular value decomposition. Eigenvector with eigenvalue approximately zero confirms the surface normal $\vec{N}_{V}$. Eigenvectors with eigenvalues $e_{1} \geq e_{2}$
estimated as directional principal direction $\vec{T}_1$ and $\vec{T}_2$ at vertices $V$.

\textit{Step 04-Estimating Principal Curvatures $k_{1}^{V}$, $k_{2}^{V}$:}

Now, for vertices $V$, we have principal directions $\vec{T}_1$ and $\vec{T}_2$ from previous step. Let, $\theta_{i}$ be the angle between any tangent direction $\vec{T}_{i}$ and principal direction $\vec{T}_1$. Similar to the continuous case, we obtain the following

\begin{align}
e_{1} &= \vec{T}_{1}^{T}M_{V}\vec{T}_{1} 
      =A \cdot k_{1}^{V} + B \cdot k_{2}^{V} .
\end{align}
\begin{align}
e_{2} &= \vec{T}_{2}^{T}M_{V}\vec{T}_{2} 
      = B \cdot k_{1}^{V} + C \cdot k_{2}^{V} .
\end{align}

where,
$ A=\sum_{{V}_{i} \in \tilde{V}}w_{i}cos^{4}\theta_{i}$ , \\
$B =\sum_{{V}_{i} \in \tilde{V}}w_{i}\sin^{2}\theta_{i}\cos^{2}\theta_{i} $, $C=\sum_{{V}_{i} \in \tilde{V}}w_{i}\sin^{4}\theta_{i}$.\\

\begin{algorithm}
\caption{Computational T-algorithm for Darboux Frame Features Extraction for Each Subject} 
\begin{algorithmic}
    \State \textbf{Input:} 3D sMRI, Segmented WM and GM 
    \State \textbf{Output:} Darboux frame-based features
    \State Surface Generation: Surface $S$ with vertices $V$, neighbors $V_{i}$ with 3-D locations ($x_i,y_i,z_i$) in a triangular mesh with estimated Surface Normal following Eq.10
    
    \For{$k = 1 : \text{len}(\text{vertices})$}
        \State Compute Tangent Direction following Eq.11
        \State Compute Directional Curvature following Eq.7
        \State Construct Curvature Matrix $M_{V}$ following Eq.17,weight $w_{i}$ using Eq.18
        \State Singular Value Decomposition of $M_{V}$ to get principal directions $\vec{T}_{1},\vec{T}_{2}$
        \State Compute principal curvatures $k_{1}^{V}$, $k_{2}^{V}$ using Eq.(19),(20)
        \State Compute Mean Curvature $H=\frac{k_{1}^{V}+k_{2}^{V}}{2}$, Gaussian Curvature $K_G=k_{1}^{V} \cdot k_{2}^{V}$
        \State Save $\vec{T}_{1},\vec{T}_{2}, \vec{N},k_{1}^{V},k_{2}^{V},H,K_G$
    \EndFor
    
\end{algorithmic}
\end{algorithm}
\vspace{-0.8cm}
\section{Fusion of disentangled semantic representations  with DFF }
Variations in MRI scanners, imaging protocols, and patient populations change data distributions \cite{bib26}. Consequently, images from multiple sites lead to domain shift.  However, BMI studies rarely explicitly tackle multi-site shifts, revealing a significant research gap. These shifts can be divided into covariate and concept shifts: covariate shift results from differences in MRI scanners, while concept shift arises from variations in patient populations; for example, one site may have fewer cases of obesity (BMI $\geq$30) than others. Therefore, the proposed framework addresses covariate and concept shifts by leveraging DSR\cite{bib26, bib28, bib29} learning and label distribution-aware margin (LDAM) loss\cite{bib30}, respectively. 
\vspace{-0.3cm}
\subsection{Reconstruction of Semantic Latent Variables}
Given, image $x$, generated from two independent latent variables: $z_s$ (site) and $z_y$ (semantic), $z=[z_s,z_y]$. 
The reconstruction component extracts features using a backbone feature extractor, such as ResNet-18, and then employs a variational auto-encoder (VAE) to reconstruct the feature. It encodes the feature into latent variables $z$ and uses these to reconstruct $G(x)$ through two independent latent variables, $z_s$ and $z_y$. Considering a semantic encoder $H_y$ and a site encoder $H_s$, the variational lower bound of the marginal likelihood is presented as follows

\begin {align}
\mathcal{L}_{ELBO} (\phi_{y},\phi_{s}, \theta_{r}) = -D_{KL} (q_{\phi_{y}} (z_y |G(x) || P(z_y)) \nonumber \\
- D_{KL} (q_{\phi_{s}} (z_s |G(x) || P(z_s)) \nonumber \\
+ \mathbb{E}_{\phi_{y,s}}(z_y,z_s |G(x)) [\log P_{\theta_{r}} (G(x)|z_y,z_s)] ,
\label{eq:elbo}
\end{align}

where the first term in Equation (\ref{eq:elbo}) is the Kullback--Leibler (KL) divergence regulariser for the semantic encoder $H_y$. It pushes the posterior $q_{\phi_{y}} (z_y |G(x)$ to stay close to the prior $N(0,1)$. 
 The second term is the identical regulariser for site encoder $H_S$. Finally, the third component is the reconstruction log-likelihood.
\vspace{-0.3cm}
\subsection{Disentanglement of Semantic Latent Variables}
The disentanglement component includes a semantic adversarial learning module and a site adversarial learning module. With the semantic latent variables $z_y$, the semantic adversarial module combines semantic information while removing site information. This is acheived via a semantic classifier $C_y$ and site classifier $C_s$. Gradient reversal layer (GRL) is used for $C_s$ to exclude site information. Additionally, to address concept shift, the semantic classification loss includes a LDAM loss .

\begin{align}
\mathcal{L}_{sem} (\phi_{y},\theta_{y,y}, \theta_{y,s}) = \nonumber \\
\frac{\delta}{n_s} \sum \limits_{x_i^{s_1} \in D_{s_1}} \mathcal{L}_{y} ( C_y (H_y(G(x);\phi_y);\theta_{y,y}),y_i) \nonumber \\
- \frac{\lambda}{n} \sum \limits_{x_i \in (D_{s_1},D_{s_2},D_{s_3})} \mathcal{L}_{s} ( C_s (H_s(G(x);\phi_y);\theta_{y,s}),s_i) .
\end{align}

\begin{align}
\mathcal{L}_{y} ((z_y,y_i);C_y)= -\log\frac{e^{l_{y_i}-\triangle_{y_i}}}{e^{l_{y_i}-\triangle_{y_i}}+\sum \limits_{j\neq y_i}e^{l_j}} \nonumber \\
where, \triangle_{y_i}= \frac {C} {n_j^{1/4}} for j \in {1,...k} .
\end{align}
In equation (22), $\lambda$ is the trade-off parameter, and a bigger $\delta$ enables learning more semantic information.
Imbalanced data across sites causes concept shift. Assuming the test label distribution is known, the LDAM loss assigns larger margins to the minority class \cite{bib30}.In equation (23), the ${l_{y_i}}$ is the logit score of sample $i$. $C$ is the hyperparameter to be tuned and the margin $\triangle_{y_i}$ depends on the label distribution. Minority classes (small $n_j$ samples) get a larger margin by ${n_j^{1/4}}$.
The objective function for the site adversarial learning module is defined as follows:

\begin{align}
\mathcal{L}_{site} (\phi_{s},\theta_{s,s}, \theta_{s,y}) = \nonumber \\
\frac{1}{n} \sum \limits_{x_i \in (D_{s_1},D_{s_2},D_{s_3})} \mathcal{L}_{s} ( C_s (H_s(G(x);\phi_s);\theta_{s,s}),s_i) \nonumber \\
- \frac{\omega}{n} \sum \limits_{x_i \in (D_{s_1},D_{s_2},D_{s_3})} \mathcal{L}_{E} ( C_y (H_s(G(x);\phi_s);\theta_{s,y})) .
\end{align}
where, $\omega$ is the trade-off parameter. The overall loss function for DSR is as follows:

\begin{align}
\mathcal{L}_{DSR} (\phi_y,\theta_{y,s},\theta_{y,y},\phi_{s},\theta_{s,s},\theta_{s,y},\theta_r) = \nonumber \\ \mathcal{L}_{ELBO} +\beta\mathcal{L}_{sem} +\gamma\mathcal{L}_{site} .
\end{align}
\vspace{-0.8cm}
\subsection{Fusion of DSR with DFF}
Given semantic latent representation $z_y$ from DSR and the Darboux Frame-based tabular features $h_t$, the logits are $l_{z_y}=W_yz_y$ and $l_t=W_th_t$ with trained model weights $W_y$ and $W_t$, respectively; the gated fusion of logits is the following:

\begin{align}
l_{fusion} = g_{DSR}\odot l_{z_y} + g_t\odot l_t; \nonumber \\where,  g_{DSR} +g_t=1 .
\end{align}

Therefore, the label loss on the fused logit becomes the following:
\begin{align}
\mathcal{L}_{y} ((z_y,y_i);C_y)= -\log\frac{e^{l_{fusion,y_i}-\triangle_{y_i}}}{e^{l_{fusion,y_i}-\triangle_{y_i}}+\sum \limits_{j\neq y_i}e^{l_{fusion,j}}} .
\end{align}
To reduce redundancy between DSR and Darboux feature branches, the orthogonal loss $\mathcal{L}_{ortho}$ with $\hat{z_y}=z_y/||z_y||_2$ and $\hat{h_t}=h_t/||h_t||_2$ with per-sample weight $s_i$, the per-sample squared cosine similarity as follows:
\begin{align}
\mathcal{L}_{ortho} = \frac {\sum\limits_{i}s_i(\hat{z_{i,y}^T}\hat{h_{t,i}})^2}{\sum \limits_{i} s_i} .
\end{align}
To keep the fusion prediction close to the DSR prediction, the KL anchor loss with $p_{z_{y}}=softmax(l_{z_y})$ and $p_{fusion} = softmax(l_{fusion})$, the loss function $\mathcal{L}_{KL}$ is as follows:

\begin{align}
\mathcal{L}_{KL} = KL(p_{z_{y}}||p_{fusion} ) .
\end{align}
With weights $\lambda_{ortho}(e)$ and $\lambda_{KL}(e)$  ramped linearly with each epoch $e$ the fusion loss is as follows:

\begin{align}
\mathcal{L}_{fusion} = \lambda_{ortho}(e)\mathcal{L}_{ortho} + \lambda_{KL}(e)\mathcal{L}_{KL} .
\end{align}
Finally, the combined loss of DSR with the Darboux frame feature is as follows:
\begin{align}
\mathcal{L}_{combine}= \mathcal{L}_{DSR} + \mathcal{L}_{fusion}+\lambda_{y}\mathcal{L}_{y} .
\end{align}

\section{Decoupled Uncertainty Quantification for Multi-site Data}
Uncertainty indicates the trustworthiness (validity) of DL model probability estimates\cite{bib31}. Evidential DL (EDL) is a method for uncertainty quantification that extracts features, called `evidence', from images.  EDL models the predictive uncertainty using a Dirichlet distribution over class probabilities \cite{bib32}. Aleatoric uncertainty reflects inherent data noise or label ambiguity, while epistemic uncertainty arises from limited data or insufficient understanding of the data-generating process.\\
With $K$ classes, the sample space is $X$, and the label space is $Y$. The probability vector $\textbf{p}=(p_1,..,p_k) \in \mathbb{P}_1(Y)$. The probability simplex $\mathbb{P}_1(Y)$ over the label space, i.e. $\mathbb{P}_1(Y) = \{ \textbf{p} \in [0,1]^K | \sum \limits_{j=1}^K p_j=1 \}$.
To train a deep learning model predicting Dirichlet distribution parameters with a non-informative prior $a=[1,..1]$, the model's output $\phi(x)$ goes through a non-negative activation function $\sigma(.)$, such as ReLU, to produce the evidence vector $	\textbf{e}=\lbrack e_1,..e_k\rbrack$, where $\textbf{e}=\sigma(\phi(x))$.The estimated Dirichlet parameter $\alpha=e+a$. The Dirichlet distribution $Dir(\alpha)$ over categorical distribution \textbf{p} is a second-order distribution with $S_o=\sum \limits_{j=1}^K \alpha_j$. The expectation $\mathbb{E}[\textbf{p}]$ indicates the aleatoric uncertainty, and the concentration parameter $S_o$ determines the dispersion around the mean. Low $S_o$ indicates high uncertainty, while high $S_o$ shows high confidence.
\vspace{-0.3cm}
\subsection{Vicinal Loss for Aleatoric Uncertainty Estimation}
With randomly selected paired training samples $(x^{(m)},y^{(m)})$ ,  $(x^{(n)},y^{(n)})$ and $\lambda_{AU} \in [0,1]$ sampled from beta distribution $Beta(\beta,\beta)$ where $\beta>>1$ (e.g. $\beta=10$). The interpolated training samples with the selected training samples:
\begin{align}
\tilde{x} =\lambda_{AU}  x^{(m)} + (1-\lambda_{AU} ) x^{(n)} \nonumber \\
\tilde{y} =\lambda_{AU}  y^{(m)} + (1-\lambda_{AU} ) y^{(n)} .
\end{align}
The loss of deep evidential learning (EDL) is is presented as:
\begin{align}
\mathcal{L}_{1} (\textbf{p},\tilde y) = \sum \limits_{j=1}^K \tilde{y}_j (\psi(S) -\psi(\alpha_j)),
\end{align}
where $\psi(.)$ is the digamma function. With vicinal pair $(\tilde{x},\tilde{y})$, the vicinal objective function is formulated as:

\begin{align}
\mathcal{L}_{vicinal} = \mathbb{E}_{((x^{(m)},y^{(m)}),(x^{(n)},y^{(n)}))\sim D)} \nonumber \\ \mathbb{E}_{\lambda\sim Beta(\beta,\beta)} \mathbb{E}_{\textbf{p}\sim Dir(\tilde\alpha|\tilde {x})} [\mathcal{L}_{1} (\textbf{p},\tilde y)] .
\end{align}

\subsection {Noise-Augmented Loss for Epistemic Uncertainty Estimation} Mixing training sample $(x^{(n)},y^{(n)})$ with noisy sample 
$x^{(m)}$ sampled from gaussian noise $(x^{(m)} \sim \mathcal{N} (0,\sigma^2)$ and label $y^{(m)}$ as a uniform distribution

\begin{align}
\tilde{x} =\lambda_{EU} x^{(m)} + (1-\lambda_{EU}) x^{(n)} \nonumber \\
\tilde{y} =\lambda_{EU} y^{(n)} + (1-\lambda_{EU})[\frac{1} .{K},...,\frac{1}{K}]
\end{align}

where $\lambda_{EU} \sim Beta(\beta^+_{noise},\beta^-_{noise})$ with $\beta^+_{noise}>>\beta^-_{noise}$ ensures the original samples' dominance without degradation of predictive performance. The noise-augmented loss function is presented as:

\begin{align}
\mathcal{L}_{noise} = \mathbb{E}_{((x^{(n)},y^{(n)})\sim D,x^{(m)} \sim \mathcal{N} (0,\sigma^2))}  \nonumber \\ \mathbb{E}_{\lambda_{EU}\sim Beta(\beta^+_{noise},\beta^-_{noise}) } \mathbb{E}_{\textbf{p} \sim Dir(\tilde{\alpha}|\tilde{x})} [\mathcal{L}_{1} (\textbf{p},\tilde y)] .
\end{align}
Finally, the total vicinal loss to jointly optimize the model for training data fit and predictive uncertainty is defined as:

\begin{align}
\mathcal{L}_{UQ} = \mathcal{L}_{vicinal} + \mathcal{L}_{noise} .
\end{align}

\section{Experiments}
\subsection{Experimental Dataset}
We evaluated the framework using multisite brain imaging datasets from the Human Connectome Project (HCP) \cite{bib34}, Cambridge Centre for Ageing and Neuroscience (Cam-CAN) \cite{bib35}, and Information Extraction from Images (IXI) \cite{bib36}. 
The three data sets contain T1-weighted MRI sequences, scanner names, and physiological and demographic data (BMI, age, and sex). Scanners include the Siemens 3T Connectome Skyra, Siemens 3T TIM Trio, and Philips Medical Systems Intera 3T, each with different parameters.
Exclusion criteria were non-right-handedness, diagnosis upon drug addition, and alcohol dependence or abuse (positive breath or urine test). The IRB approved the dataset under protocol \#2434358. 
Table 1 shows subject distributions across the training, validation, and test splits for three datasets. Raw T1-weighted MRI scans were preprocessed with an automated FSL pipeline \cite{bib37}, including skull stripping, normalization, and nonlinear transformation to MNI152 space \cite{bib38}. To reduce computational demands, scans were resampled to 90$\times$90$\times$90 voxels at 1 mm isotropic resolution.

\begin{table}[!t]
\caption{Class $\times$ Site (Dataset) Sample Counts for the Train, Validation, and Test Splits}
\label{tab:split_counts}
\centering
\footnotesize
\centering
\centering
\scriptsize
\setlength{\tabcolsep}{3.5pt}
\renewcommand{\arraystretch}{1.15}
\begin{tabular}{l c c c c}
\toprule
BMI Class & CamCAN & HCP & IXI & Total \\
\midrule
\multicolumn{5}{l}{\textit{Training}} \\
\hspace{1em}BMI $<25$ & 198 & 218 & 209 & 625 \\
\hspace{1em}BMI $\ge 30$ &  67 & 107 &  43 & 217 \\
\hspace{1em}\textbf{Total} & \textbf{265} & \textbf{325} & \textbf{252} & \textbf{842} \\
\midrule
\multicolumn{5}{l}{\textit{Validation}} \\
\hspace{1em}BMI $<25$ & 42 & 47 & 45 & 134 \\
\hspace{1em}BMI $\ge 30$ & 15 & 23 &  9 &  47 \\
\hspace{1em}\textbf{Total} & \textbf{57} & \textbf{70} & \textbf{54} & \textbf{181} \\
\midrule
\multicolumn{5}{l}{\textit{Testing}} \\
\hspace{1em}BMI $<25$ & 43 & 47 & 46 & 136 \\
\hspace{1em}BMI $\ge 30$ & 14 & 23 &  9 &  46 \\
\hspace{1em}\textbf{Total} & \textbf{57} & \textbf{70} & \textbf{55} & \textbf{182} \\
\bottomrule
\end{tabular}
\end{table}
\vspace{-0.5cm}
\subsection{Computational Framework Implementation Details}
We trained and tested our proposed method (Fig. \ref{fig:wide_figure}) using the TensorFlow 2.16.1 framework on an NVIDIA Tesla V100 SXM2 GPU with 16 GB of memory. For each patient, to derive brain-region-specific, Darboux-frame-based features, we applied AAL3 atlas-based parcellation \cite{bib39} to map each vertex coordinate and its features from the left and right hemisphere 3D meshes for each subject. 
The principal and normal direction values are converted to norm values. 
The fusion part includes subject age, sex, and site-invariant significant features.
The orthogonal and KL loss weights [$\lambda_{ortho}(e),\lambda_{KL}(e)$] during fusion are initialized to 0.015 and 0.05, with linear warmups at epochs 50 and 30, respectively. The weight ($\lambda_{y}$) for the LDAM loss is 0.5, with margins of 0.3838 for BMI$<$25 and 0.5000 for BMI$\geq$30, respectively. The dimensions of the disentangled semantic and site latent spaces are 128 and 64, respectively. The weights ($\beta, \gamma$) for semantic and site loss are initialized at 0.1. The model is trained for 300 epochs, with the learning rate initialized at 3e-4 and 8 samples per BMI group in each batch. We evaluate the performance of different models using accuracy (Acc.), precision (Prec.), recall (Rec.), and F1-score (F1) across the training, validation, and testing distributions. To compare uncertainty quantification methods across modeling techniques, we use excess area under the risk--coverage curve (E-AURC $\times$ 1000; lower is better). Lower E-AURC indicates more reliable aleatoric uncertainty estimates and better predictive performance on leave-one-site-out (LOSO) data \cite{bib33}.

\section{Results \& Discussion}
\subsection{Predictive Performance}
In our experiment, we evaluate different modeling techniques: loss-function reweighting with focal loss, LDAM loss alone, and DSR with LDAM. Each technique is tested with a ResNet-18 architecture, with and without DFF. We also evaluate DSR with LDAM with and without geometric mean (G-mean) during optimization. The results are shown in \textbf{Table II}. On the test data, the best performance is achieved by combining DSR, DFF, and G-mean. The addition of DFF improves test results, showing that cortical folding direction provides information beyond voxel-level features. The overall model (DSR+LDAM + DFF with G-mean) achieves the highest test recall (0.785) and F1 (0.747), which are the most relevant metrics under imbalanced BMI groups. While some baselines outperform our model on the validation split, our full model consistently achieves the best recall and F1 on the testing, the split that best reflects performance on unseen data.

\begin{table*}[!t]
\renewcommand{\arraystretch}{1.3}
\caption{Predictive Performance Comparison of Modeling Techniques. The Best result is shown in Bold for each performance metric.}
\label{tab:model_performance}
\centering
\centering
\centering
\scriptsize
\setlength{\tabcolsep}{3.5pt}
\renewcommand{\arraystretch}{1.2}
\begin{tabular}{l cccc cccc cccc}
\toprule
\multirow{2}{*}{\textbf{Modeling Techniques}} &
\multicolumn{4}{c}{\textbf{Training}} &
\multicolumn{4}{c}{\textbf{Validation}} &
\multicolumn{4}{c}{\textbf{Testing}} \\
\cmidrule(lr){2-5} \cmidrule(lr){6-9} \cmidrule(lr){10-13}
 & Acc. & Prec. & Rec. & F1
 & Acc. & Prec. & Rec. & F1
 & Acc. & Prec. & Rec. & F1 \\
\midrule
 Focal loss with ResNet18 & 0.6900 & 0.6158 & 0.6303 & 0.6202 & 0.7017  & 0.6195  & 0.6258 & 0.6222  & 0.6484  & 0.5573 & 0.5633  & 0.5588 \\
LDAM loss with ResNet18 & 0.7233 & 0.6968 & 0.7534 &  0.6953 & 0.6851 & 0.6687  & 0.7182 & 0.6591 & 0.6923  & 0.6648 & 0.7150 &  0.6601\\
LDAM loss with DSR & 0.8242  & 0.7740 & 0.8184 & 0.7890 & 0.7293 & 0.6738 & 0.7067 & 0.6817 &0.7692  & 0.7021 & 0.7233 & 0.7105 \\
Focal-loss + ResNet18 + DFF & 0.8040 & 0.7441 & 0.7492 & 0.7466 & 0.7238 & 0.6377 & 0.6339 & 0.6357 & 0.7473 & 0.6562 & 0.6367 & 0.6440 \\
LDAM + ResNet18 + DFF & 0.8207 &0.7734 &0.7303 &0.7467 & \textbf{0.7845} & 0.7202 & \textbf{0.7232} & \textbf{0.7217} & \textbf{0.7912} &0.7228 &0.6949 & 0.7060 \\
DSR + LDAM + DFF \\(without G-mean) & 0.8147 & \textbf{0.8067} & 0.6721 & 0.6991 & 0.7735 & \textbf{0.7260} & 0.6053 & 0.6156 & \textbf{0.7912} & \textbf{0.7440} & 0.6393 & 0.6564 \\
DSR + LDAM + DFF \\(with G-mean) & \textbf{0.8325} & 0.7878 & \textbf{0.8481} & \textbf{0.8045} & 0.7017 & 0.6572 &0.6949 & 0.6605 & 0.7857 & 0.7342 & \textbf{0.7847} & \textbf{0.7472} \\
\bottomrule
\end{tabular}
\end{table*}
\vspace{-0.3cm}
\subsection{Comparison}
\subsubsection{Feature Fusion}
We compare modeling performance using different features instead of DFF. For example, we test cortical thickness, gray matter volume, white matter volume, and only Gaussian curvature. \textbf{Table III} presents a comparison in which the DFF branch is replaced by other features. Regarding validation samples, the performance of the model using DFF, cortical thickness, and Gaussian curvature falls within a similar range, and the underlying reasoning may be the relationship between cortical thickness and curvature. Higher BMI is associated with lower gray matter volume and cortical thickness, but greater curvature, linking BMI to cortical morphology. 

\subsubsection{Modeling Techniques with Uncertainty Quantification}
We compare our proposed method with existing techniques in the literature and quantify performance using E-AURC with decoupled uncertainty quantification. The results are presented in \textbf{Table IV}. Based on E-AURC, our proposed method has the lowest value, followed by the model without DFF, which has an E-AURC value of 116.522. Fig.3 shows aleatoric and epistemic uncertainty estimates for correct and incorrect predictions in the test data across acquisition sites. It is observed in both cases (with and without DFF) that the aleatoric uncertainty (AU) for correct predictions has a lower median value compared to incorrect predictions across all acquisition sites. However, for epistemic uncertainty (EU), the proposed model with DFF (Fig. 3(b)) produces more negative (more confident) values for correct predictions across all sites. In contrast, without DFF (Fig. 3(d)), incorrect predictions, particularly for HCP and IXI, are more negative.
\begin{table*}[!t]
\renewcommand{\arraystretch}{1.3}
\caption {Comparison of Feature Fusion Modeling Techniques. The Best result is shown in Bold for each performance metric.}
\label{tab:fusion_comparison}
\centering
\centering
\centering
\scriptsize
\setlength{\tabcolsep}{3.5pt}
\renewcommand{\arraystretch}{1.2}
\begin{tabular}{l cccc cccc cccc}
\toprule
\multirow{2}{*}{\textbf{Modeling Techniques}} &
\multicolumn{4}{c}{\textbf{Training}} &
\multicolumn{4}{c}{\textbf{Validation}} &
\multicolumn{4}{c}{\textbf{Testing}} \\
\cmidrule(lr){2-5} \cmidrule(lr){6-9} \cmidrule(lr){10-13}
 & Acc. & Prec. & Rec. & F1
 & Acc. & Prec. & Rec. & F1
 & Acc. & Prec. & Rec. & F1 \\
\midrule
DSR + LDAM \\+ Cortical Thickness & 0.7577 &0.6936 &0.7165 &0.7020 &0.7072 &0.6571 &0.6917 & \textbf{0.6623} &0.7527 &0.6811 &0.6979 &0.6878 \\
DSR + LDAM \\+ Gray Matter Volume &0.2910 &0.4628 &0.4833 &0.2717 &0.3149 &0.5944 &0.5304 &0.2883 &0.3022 &0.4942 &0.4971 &0.2865 \\
DSR + LDAM \\+ White Matter Volume &0.7399 &0.6120 &0.5240 &0.4907 &\textbf{0.7238} &0.5388 &0.5095 &0.4719 &0.7308 &0.5590 &0.5177 &0.4904 \\
DSR + LDAM \\+ Gaussian Curvature &0.7447 &0.6872 &0.7137 &0.6931 &0.6961 &0.6414 &0.6705 &0.6462 &0.7637 &0.6922 &0.7052 &0.6978 \\
 DSR + LDAM \\+ DFF  & \textbf{0.8325} & \textbf{0.7878} & \textbf{0.8481} & \textbf{0.8045} & 0.7017 & \textbf{0.6572} & \textbf{0.6949} & 0.6605 & \textbf{0.7857} & \textbf{0.7342} & \textbf{0.7847} & \textbf{0.7472} \\

\bottomrule
\end{tabular}
\end{table*}

\begin{table*}[!t]
\renewcommand{\arraystretch}{1.3}
\caption{Performance Comparison of the Proposed Method [DSR + LDAM + DFF + UQ] with methods in the literature and Uncertainty Quantification.The Best result is shown in Bold for each performance metric. }
\label{tab:model_performance_eaurc}
\centering
\centering
\scriptsize
\setlength{\tabcolsep}{3.5pt}
\renewcommand{\arraystretch}{1.2}
\begin{tabular}{p{2.3cm} cccc cccc cccc c}
\toprule
\multirow{2}{*}{\textbf{Modeling Techniques}} &
\multicolumn{4}{c}{\textbf{Training}} &
\multicolumn{4}{c}{\textbf{Validation}} &
\multicolumn{4}{c}{\textbf{Testing}} &
\multirow{2}{*}{\textbf{E-AURC $\Downarrow$}} \\
\cmidrule(lr){2-5} \cmidrule(lr){6-9} \cmidrule(lr){10-13}
 & Acc. & Prec. & Rec. & F1
 & Acc. & Prec. & Rec. & F1
 & Acc. & Prec. & Rec. & F1
 & \\
\midrule
 DSR + LDAM \\ +UQ  & 0.8242 &0.7740 &0.8184 &0.7890 & \textbf{0.7293} & \textbf{0.6738} & \textbf{0.7067} & \textbf{0.6817} &0.7692 &0.7021 &0.7233 &0.7105 &116.522 \\
Deep Learning (DL) \\ +UQ \cite{bib21}  &0.6116 &0.5296 &0.5338 &0.5288 &0.6243 &0.5463 &0.5529 &0.5464 &0.5989 &0.5136 &0.5158 &0.5115 &182.831 \\
DL Neural \\Network(NN)+UQ \cite{bib22} &0.6983 &0.6441 &0.6750 &0.6489 &0.6740 &0.6218 &0.6486 &0.6240 &0.7143 &0.6583 &0.6937 &0.6647 & 409.387 \\
Domain Adversarial NN +UQ \cite{bib40}  &0.5202 &0.4440 &0.4331 &0.4344 &0.5249 &0.4469 &0.4374 &0.4385 &0.6099 &0.5080 &0.5088 &0.5075 &216.785 \\
Proposed Method & \textbf{0.8325} & \textbf{0.7878} & \textbf{0.8481} & \textbf{0.8045} & 0.7017 & 0.6572 &0.6949 & 0.6605 & \textbf{0.7857} & \textbf{0.7342} & \textbf{0.7847} & \textbf{0.7472} & \textbf{67.962}\\
\bottomrule
\end{tabular}
\end{table*}

\begin{figure}[!t] \label {fig: uncertainity quantification}
\centering
\includegraphics[width=0.3\textwidth]{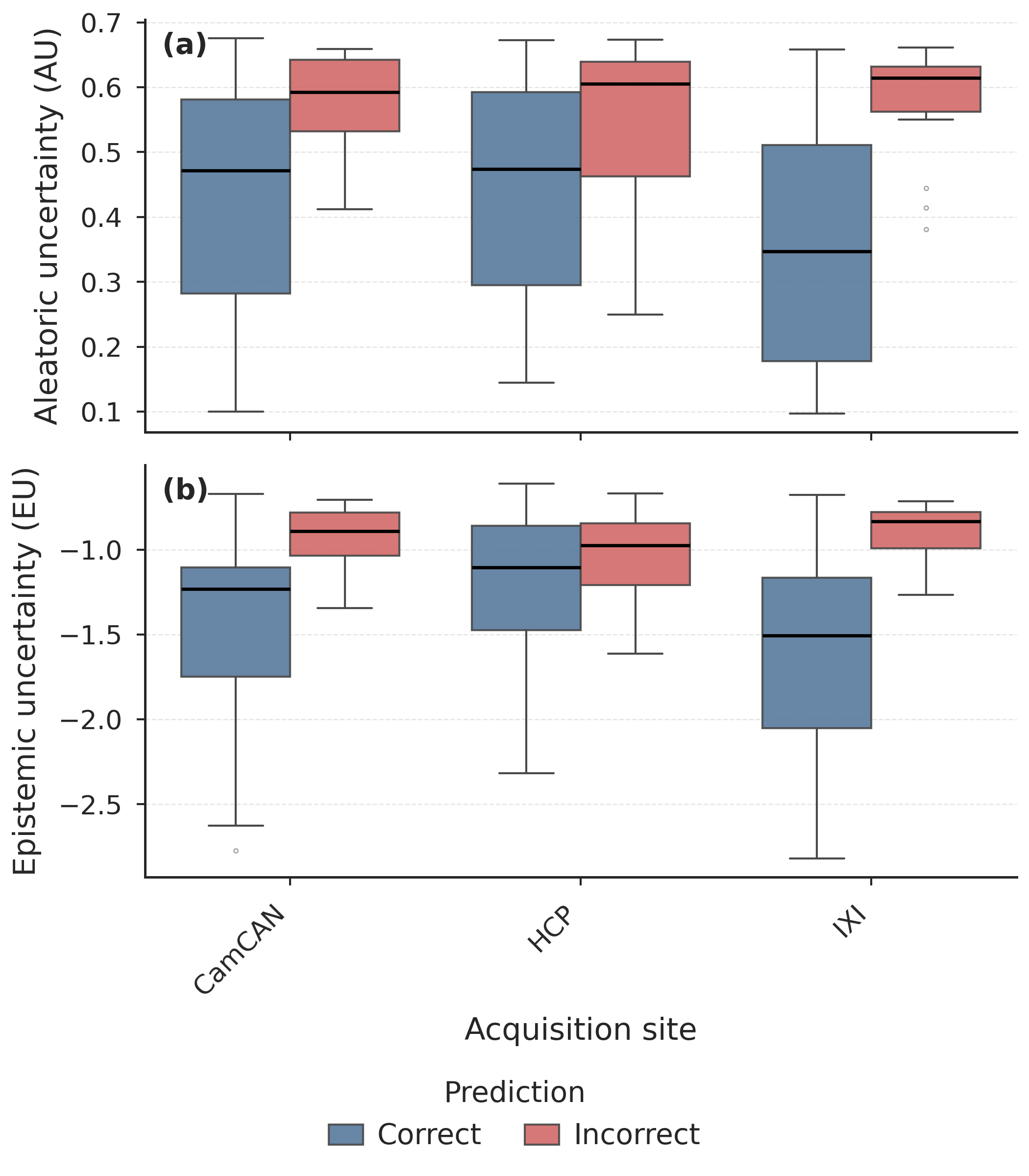}\\[1pt]
\includegraphics[width=0.3\textwidth]{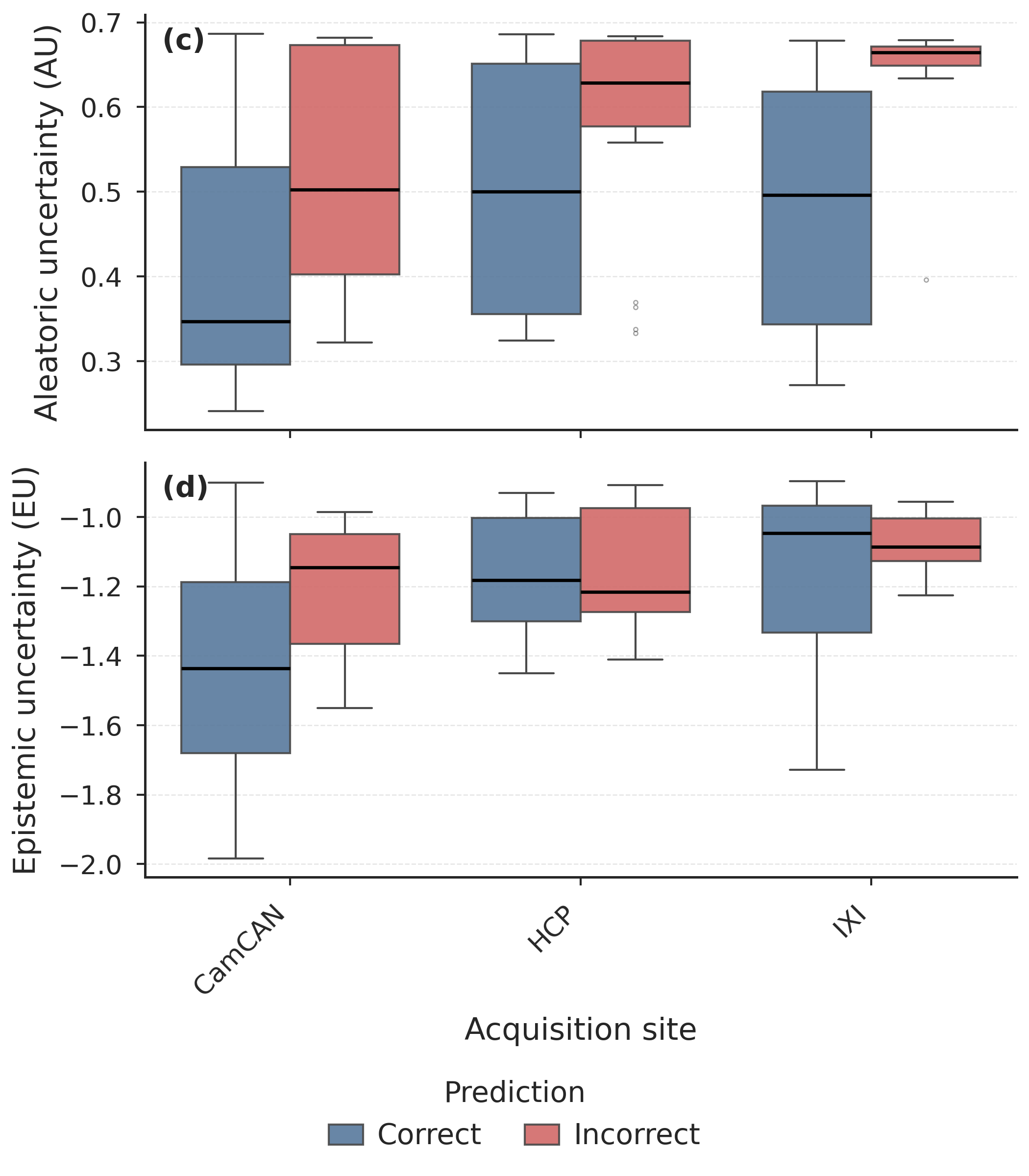}\\[1pt]
\caption{Uncertainty by acquisition site for correctly (blue) and incorrectly (red)
classified test samples. (a)~AU and (b)~EU for \emph{Proposed Method with Darboux Feature Fusion (DFF)}; (c)~AU and (d)~EU for
\emph{Without DFF}. Each box represents the interquartile range, with the median marked by a black line.}
\label{fig:uncertainty_all}
\end{figure}
\vspace{-0.1cm}
\subsection{Generalizability of Multi-site Uncertainty Quantification via  LOSO Analysis}
To evaluate the uncertainty quantification (UQ) performance of the proposed method, we compare it with methods from the literature on the BMI prediction task using LOSO validation. UQ is evaluated using E-AURC on the test samples. For each evaluation, the test set comprises three different sites. The proposed method achieves a lower E-AURC across all three test sites. Moreover, to assess the importance of DFF, we also evaluate the model without DFF (DSR + LDAM + UQ), with the results presented in \textbf{Table V}. Across sites with different MRI scanners, acquisition parameters, BMI groups, or label distributions, the proposed model more reliably signals uncertainty in its predictions.

\subsection {Brain Region Identification}
\begin{figure}[!t] \label {fig: identified brain regions BMI>=30}
\centering
\includegraphics[width=0.5\textwidth]{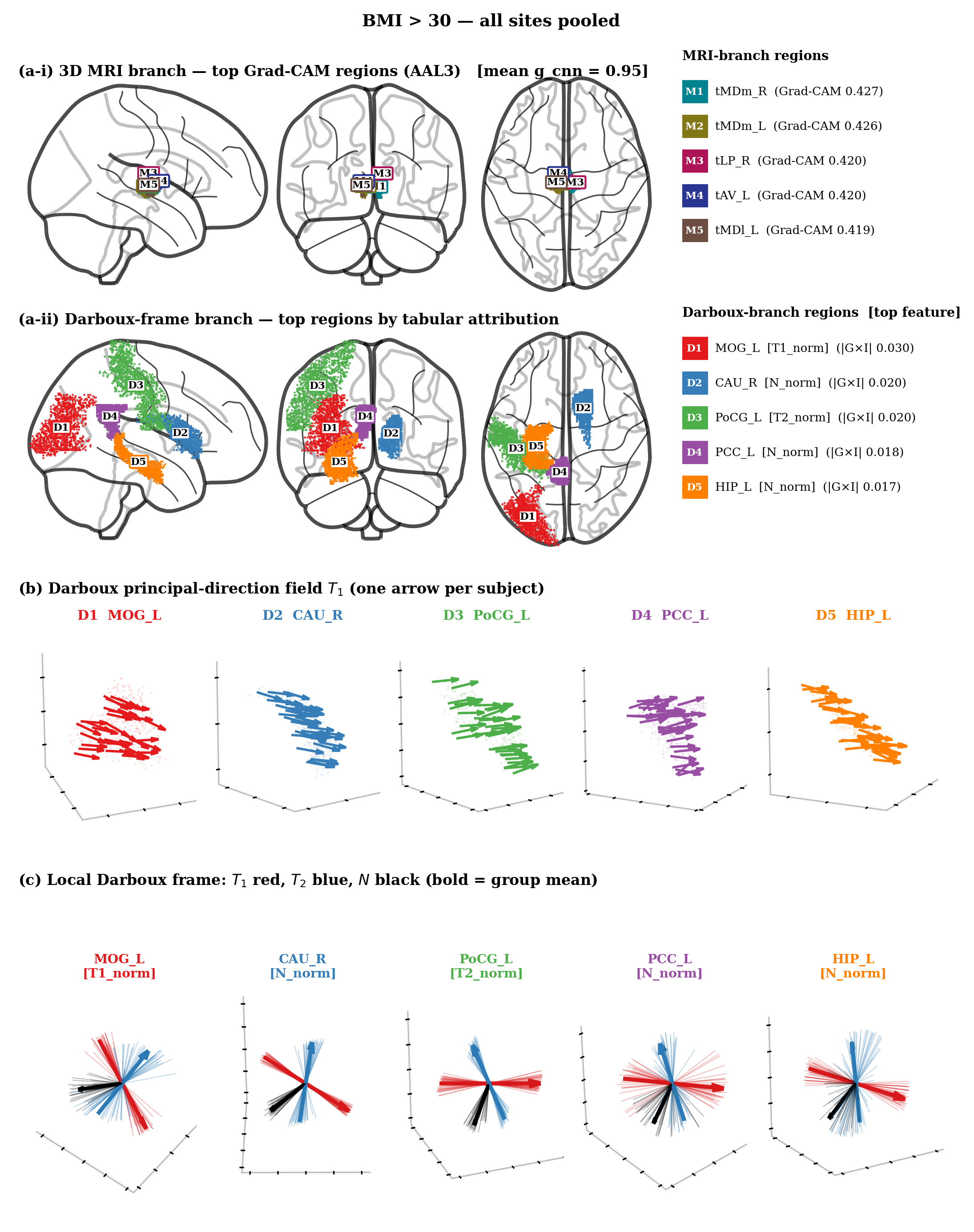}\\[3pt]
\caption{Top 5 brain regions identified by the proposed method for correctly predicted test samples with quantified epistemic uncertainty across all sites (HCP, IXI, and CamCAN) for \textbf{BMI $\ge$30} groups. (a-i) 3D-MRI-based branch of the proposed method identified brain regions ($t\mathrm{MDm\_L}$, $t\mathrm{MDm\_R}$: left and right thalamus mediodorsal medial magnocellular subcortical regions; $t\mathrm{LP\_R}$: right thalamus lateral posterior subcortical region; $t\mathrm{AV\_L}$: left thalamus anterpventral Nucles subcortical region; $t\mathrm{MDl\_L}$: left thalamus mediodorsal lateral parvocellular subcortical region ) during prediction; (a-ii) Darboux-frame branch of the proposed method identified brain regions ( $\mathrm{MOG\_L}$ : left middle occipital gyrus under occipital lobe; $\mathrm{CAU\_R}$: right caudate subcortical regions ;$\mathrm{PoCG\_L}$ : left postcentral gyrus under sensorimotor ; $\mathrm{PCC\_L}$ : left posterior cingulate gyrus under cingulate cortex ; $\mathrm{HIP\_L}$: left hippocampus under subcortical regions )during prediction ; (b) Darboux-frame principal tangent direction $T_1$ in the top 5 identified brain regions , one arrow per subject or sample; (c) Local Darboux-frame ($T_1 (red) , T_2 (blue), N(black)$), the group mean across all corrected predictied subjects shown in bold.}
\label{fig:brain_regions}
\end{figure}

To identify regions driving the predictions, we applied gradient-based localization mapping \cite{bib41}. The maps are computed per subject in MNI space ($1mm^3$), aggregated across BMI groups for corrected predicted test samples from all sites (HCP, IXI, CamCAN), and overlaid on the AAL3 atlas\cite {bib39} (166 regions). A $25^{th}$-percentile threshold retained regions with moderate consistency between subjects. Fig. \ref{fig:brain_regions} shows top five regions per branch with quantified epistemic uncertainty.\\
For the BMI $\ge$ 30 group (n=34), the branches highlight different regions. The 3D-MRI branch concentrates on the thalamus, a multisensory hub whose functional connectivity is negatively associated with obesity phenotypes\cite {bib5}. The DFF branch instead highlights the left middle occipital gyrus, right caudate, left postcentral gyrus, left posterior cingulate, and left hippocampus (Fig.\ref{fig:brain_regions}, a-ii). Except for the occipital region, which belongs to the dorsal attention network \cite{bib15}, these regions support reward processing, sensorimotor integration, and hedonic control of food intake, consistent with the competing neural system model \cite {bib4}.\\
Unlike the voxel-based MRI branch, the DFF branch also reveals cortical folding direction in the top five regions. Together with the testing results (Tables II, III, and IV) and multi-site uncertainty quantification (Table V), the proposed method's brain region identification suggests that the geometric directional features from the Darboux frame in reward-related regions carry BMI-related signals.

\begin{table*}[!t]
\caption{Performance comparison of modeling techniques for Generalizability of Uncertainty Quantification via Leave One Site Out (LOSO) Analysis}
\label{tab:results}
\centering
\scriptsize
\setlength{\tabcolsep}{3.0pt}
\renewcommand{\arraystretch}{1.1}
\begin{tabular}{l cccc cccc cccc c}
\toprule

\multirow{2}{*}{\makecell[l]{\textbf{Modeling}\\\textbf{Techniques}}}
 & \multicolumn{4}{c}{Training (CamCAN + HCP) ($N=694$)}
 & \multicolumn{4}{c}{Validation (CamCAN + HCP) ($N=150$)}
 & \multicolumn{4}{c}{\makecell{Testing (IXI) ($N=361$)\\BMI$<$25: 300; BMI$\geq$30: 61}}
 & \multirow{2}{*}{\textbf{E-AURC$\Downarrow$}} \\
\cmidrule(lr){2-5}\cmidrule(lr){6-9}\cmidrule(lr){10-13}
 & Acc. & Prec. & Rec. & F1 & Acc. & Prec. & Rec. & F1 & Acc. & Prec. & Rec. & F1 & \\
\midrule
\rowcolor{gray!20}
\textbf{Proposed Method} &0.6902 &0.6795 &0.7150 &0.6726 &0.6867 &0.6534 &0.6786 &0.6557 &0.2798 &0.5155 &0.5144 &0.2796 &231.469 \\
DSR+LDAM+UQ &0.7363 &0.6779 &0.6302 &0.6400 &0.7600 &0.7181 &0.6507 &0.6646 &0.5457 &0.4989 &0.4981 &0.4583 &503.952 \\
DL+UQ  \cite{bib21}     &0.8501 &0.8174 &0.8483 &0.8390 &0.8000 &0.7588 &0.7588 &0.7588 &0.3601 &0.5322 &0.5432 &0.3557 &392.930 \\
DLNN+UQ  \cite{bib22}   &0.5101 &0.5936 &0.5985 &0.5097 &0.5267 &0.5676 &0.5787 &0.5205 &0.2936 &0.5029 &0.5032 &0.2934 &393.110 \\

\midrule\midrule

\multirow{2}{*}{\makecell[l]{\textbf{Modeling}\\\textbf{Techniques}}}
 & \multicolumn{4}{c}{Training (CamCAN + IXI) ($N=609$)}
 & \multicolumn{4}{c}{Validation (CamCAN + IXI) ($N=131$)}
 & \multicolumn{4}{c}{\makecell{Testing (HCP) ($N=465$)\\BMI$<$25: 312; BMI$\geq$30: 153}}
 & \multirow{2}{*}{\textbf{E-AURC$\Downarrow$}} \\
\cmidrule(lr){2-5}\cmidrule(lr){6-9}\cmidrule(lr){10-13}
 & Acc. & Prec. & Rec. & F1 & Acc. & Prec. & Rec. & F1 & Acc. & Prec. & Rec. & F1 & \\
\midrule
\rowcolor{gray!20}
\textbf{Proposed Method} &0.8670 &0.8012 &0.8873 &0.8280 &0.8321 &0.7632 &0.8412 &0.7856 &0.6645 &0.4780 &0.4985 &0.4112 &241.739 \\
DSR+LDAM+UQ &0.8522 &0.7849 &0.8694 &0.8100 &0.8626 &0.7926 &0.8346 &0.8098 &0.6688 &0.5359 &0.5017 &0.4129 &344.885 \\
DL+UQ \cite{bib21}      &0.9031 &0.8425 &0.9215 &0.8710 &0.8779 &0.8134 &0.8443 &0.8271 &0.6688 &0.5505 &0.5034 &0.4188 &268.214 \\
DLNN+UQ \cite{bib22}    &0.5961 &0.6003 &0.6502 &0.5615 &0.6336 &0.6103 &0.6630 &0.5893 &0.6710 &0.3355 &0.5000 &0.4015 &263.099 \\

\midrule\midrule

\multirow{2}{*}{\makecell[l]{\textbf{Modeling}\\\textbf{Techniques}}}
 & \multicolumn{4}{c}{Training (HCP + IXI) ($N=679$)}
 & \multicolumn{4}{c}{Validation (HCP + IXI) ($N=147$)}
 & \multicolumn{4}{c}{\makecell{Testing (CamCAN) ($N=379$)\\BMI$<$25: 283; BMI$\geq$30: 96}}
 & \multirow{2}{*}{\textbf{E-AURC$\Downarrow$}} \\
\cmidrule(lr){2-5}\cmidrule(lr){6-9}\cmidrule(lr){10-13}
 & Acc. & Prec. & Rec. & F1 & Acc. & Prec. & Rec. & F1 & Acc. & Prec. & Rec. & F1 & \\
\midrule
\rowcolor{gray!20}
\textbf{Proposed Method} &0.8144 &0.7622 &0.7990 &0.7754 &0.7823 &0.7151 &0.7075 &0.7111 &0.5858 &0.5134 &0.5161 &0.5088 &217.723 \\
DSR+LDAM+UQ &0.7305 &0.6886 &0.7350 &0.6942 &0.7347 &0.6774 &0.7097 &0.6858 &0.5040 &0.5448 &0.5577 &0.4899 &388.937 \\
DL+UQ  \cite{bib21}     &0.6362 &0.5456 &0.5495 &0.5465 &0.6531 &0.5699 &0.5775 &0.5719 &0.4987 &0.5662 &0.5817 &0.4915 &450.664 \\
DLNN+UQ  \cite{bib22}   &0.4934 &0.5842 &0.5971 &0.4907 &0.5034 &0.5898 &0.6051 &0.5001 &0.4063 &0.6095 &0.5853 &0.4031 &345.348 \\

\bottomrule
\end{tabular}
\end{table*}

\section{Conclusion}
We developed an uncertainty-aware framework for multi-site BMI prediction from sMRI with two contributions: localized DFF, which encode geometric direction of cortical folding that scalar measures cannot measure. With a dynamic fusion layer, we integrated voxel-based DSR with DFF, followed by multi-site uncertainty quantification. The study is limited to three sites, and the identified brain regions are associative, not causal. Future work will evaluate larger multimodal datasets with longitudinal and clinical covariates to determine whether the brain regions identified by DFF change with weight loss or intervention.

\appendices

\section*{Acknowledgment}
The author(s) acknowledge the use of Grammarly Pro for proofreading and editing assistance, focusing on
grammar, clarity, and style.

\end{document}